\documentclass[12pt,twoside]{article}
\usepackage{latexsym,amsmath,amssymb,hyperref}
\newenvironment{keywords}{\centerline{\bf\small
Keywords}\begin{quote}\small}{\par\end{quote}\vskip 1ex}
\def\paradot#1{}
\def\nq{\hspace{-1em}}

\begin{document}

\title{\vspace{-5ex}
\vskip 2mm\bf\Large\hrule height5pt \vskip 4mm
Can Intelligence Explode?
\vskip 4mm \hrule height2pt}
\author{{\bf Marcus Hutter}\\[3mm]
\begin{minipage}{0.45\textwidth}\raggedleft
\normalsize Research School of Computer Science \\
\normalsize Australian National University
\end{minipage}
~~\&~~
\begin{minipage}{0.4\textwidth}\raggedright
\normalsize Department of Computer Science \\
\normalsize ETH Z\"urich, Switzerland
\end{minipage}
}
\date{February 2012}
\maketitle
\vspace*{-5ex}

\begin{abstract}
The technological singularity refers to a hypothetical scenario
in which technological advances virtually explode. The most
popular scenario is the creation of super-intelligent
algorithms that recursively create ever higher intelligences.
It took many decades for these ideas to spread from science
fiction to popular science magazines and finally to attract the
attention of serious philosophers. David Chalmers' (JCS 2010)
article is the first comprehensive philosophical analysis of
the singularity in a respected philosophy journal. The
motivation of my article is to augment Chalmers' and to discuss
some issues not addressed by him, in particular what it could
mean for intelligence to explode. In this course, I will (have
to) provide a more careful treatment of what intelligence
actually is, separate speed from intelligence explosion,
compare what super-intelligent participants and classical human
observers might experience and do, discuss immediate
implications for the diversity and value of life, consider
possible bounds on intelligence, and contemplate intelligences
right at the singularity.
\def\contentsname{\centering\normalsize Contents}
{\parskip=-2.7ex\tableofcontents}
\end{abstract}

\vspace*{-2ex}
\begin{keywords} \vspace*{-2ex}
singularity;
acceleration;
intelligence;
evolution;
rationality;
goal;
life;
value;
virtual;
computation;
AIXI.
\end{keywords}
\vspace*{-3ex}

\begin{quote}\it
``Within thirty years, we will have the technological
means to create superhuman intelligence.
Shortly after, the human era will be ended.'' \par
\hfill --- {\sl Vernor Vinge (1993)}
\end{quote}

\newpage
\section{Introduction}\label{sec:Intro}

The technological singularity is a hypothetical scenario
in which self-accelerating technological advances cause
infinite progress in finite time.
The most popular scenarios are an intelligence explosion
\cite{Good:65} or a speed explosion \cite{Yudkowsky:96} or a
combination of both \cite{Chalmers:10sing}.
This quite plausibly is accompanied by a radically changing
society, which will become incomprehensible to us current
humans close to and in particular at or beyond the singularity.
Still some general aspects may be predictable.

\paradot{History of the singularity idea}
Already the invention of the first four-function mechanical
calculator one-and-a-half centuries ago \cite{Thornton:1847}
inspired dreams of self-amplifying technology. With the advent
of general purpose computers and the field of artificial
intelligence half-a-century ago, some mathematicians, such as
Stanislaw Ulam \cite{Ulam:58}, I.J. Good \cite{Good:65}, Ray
Solomonoff \cite{Solomonoff:85}, and Vernor Vinge
\cite{Vinge:93} engaged in singularity thoughts.
But it was only in the last decade that the singularity idea
achieved wide-spread popularity. Ray Kurzweil popularized the
idea in two books \cite{Kurzweil:99,Kurzweil:05}, and the
Internet helped in the formation of an initially small
community discussing this idea. There are now annual
Singularity Summits approaching a thousand participants per
year, and even a Singularity Institute.

\paradot{Artificial general intelligence}
The singularity euphoria seems in part to have been triggered
by the belief that intelligent machines that possess general
intelligence on a human-level or beyond can be built within our
life time, but it is hard to tell what is cause and effect. For
instance, there is now a new conference series on Artificial
General Intelligence (AGI) as well as some whole-brain
emulation projects like Blue Brain \cite{deGaris:10,Goertzel:10}.

\paradot{Immortalists}
A loosely related set of communities which are increasing in
momentum are the ``Immortalists'' whose goal is to extend the
human life-span, ideally indefinitely. Immortality and
life-extension organizations are sprouting like mushrooms:
e.g.\ the Immortality and the Extropy Institute, the Humanity+
Association, and the Alcor Life Extension, Acceleration
Studies, Life Extension, Maximum Life, and Methusalem
Foundations.

\paradot{Paths to singularity}
There are many different potential paths toward a singularity.
Most of them seem to be based on software intelligence on
increasingly powerful hardware. Still this leaves many options,
the major ones being %
mind uploading (via brain scan) and subsequent improvement, %
knowledge-based reasoning and planning software (traditional AI research), %
artificial agents that learn from experience (the machine learning approach), %
self-evolving intelligent systems (genetic algorithms and artificial life approach), %
and the awakening of the Internet (or digital Gaia scenario).
Physical and biological limitations likely do not allow
singularities based on (non-software) physical brain
enhancement technologies such as drugs and genetic engineering.

\paradot{Considered setup}
Although many considerations in this article should be
independent of the realized path, I will assume a virtual
software society consisting of interacting rational agents
whose intelligence is high enough to construct the next
generation of more intelligent rational agents. Indeed, one of
the goals of the article is to discuss what (super)intelligence
and rationality could mean in this setup. For concreteness, the
reader may want envisage an initial virtual world like Second
Life that is similar to our current real world and inhabited by
human mind uploads.

\paradot{Motivation}
Much has been written about the singularity and David Chalmers'
article \cite{Chalmers:10sing} covers quite wide ground. I
essentially agree with all his statements, analysis, and also
share his personal opinions and beliefs. Most of his
conclusions I will adopt without repeating his arguments. The
motivation of my article is to augment Chalmers' and to discuss
some issues not addressed by him,
in particular what it could mean for intelligence to explode.
This is less obvious than it might appear, and requires a more
careful treatment of what intelligence actually is.
Chalmers cleverly circumvents a proper discussion or
definition of intelligence by arguing %
(a) there is something like intelligence,
(b) there are many cognitive capacities correlated with intelligence, %
(c) these capacities might explode, therefore %
(d) intelligence might amplify and explode.
While I mostly agree with this analysis, it does not tell us
what a society of ultra-intelligent beings might look like. For
instance, if a hyper-advanced virtual world looks like random
noise for humans watching them from the ``outside'', what does
it mean for intelligence to explode for an outside observer?
Conversely, can an explosion actually be felt from the
``inside'' if everything is sped up uniformly? If neither
insiders nor outsiders experience an intelligence explosion,
has one actually happened?

\paradot{Contents}
The paper is organized as follows:
{\em Section~\ref{sec:WS}} briefly recapitulates the most
popular arguments why to expect a singularity and why ``the
singularity is near'' \cite{Kurzweil:05}, obstacles towards a
singularity, and which choices we have.
{\em Section~\ref{sec:SFO}} describes how an outside observer
who does not participate in the singularity might experience
the singularity and the consequences he faces. This will depend
on whether the singularity is directed inwards or outwards.
{\em Section~\ref{sec:SFI}} investigates what a participant in
the singularity will experience, which is quite different from
an outsider and depends on details of the virtual society; in
particular how resources are distributed.
{\em Section~\ref{sec:SVIE}} takes a closer look at what
actually explodes when computing power is increased without
limits in finite real time. While by definition there is a
speed explosion, who, if anyone at all, perceives an
intelligence explosion/singularity depends on what is sped up.
In order to determine whether anyone perceives an intelligence
explosion, it is necessary to clarify what intelligence
actually is and what super-intelligences might do, which is
done in {\em Section~\ref{sec:WII}}.
The considered formal theory of rational intelligence allows
investigating a wide range of questions about
super-intelligences, in principle rigorously mathematically.
{\em Section~\ref{sec:IUB}} elucidates the possibility that
intelligence might be upper bounded, and whether this would
prevent an intelligence singularity.
{\em Section~\ref{sec:SI}} explains how a society right at the
edge of an intelligence singularity might be theoretically studied
with current scientific tools.
Even when setting up a virtual society in our image, there are
likely some immediate differences, e.g.\ copying and modifying
virtual structures, including virtual life, should be very easy.
{\em Section~\ref{sec:DE}} shows that this will have immediate
(i.e.\ way before the singularity) consequences on the
diversity and value of life.
{\em Section~\ref{sec:Misc}} contains some personal remarks and
{\em Section~\ref{sec:Conc}} draws some conclusions.

\paradot{Terminology}
I will use the following terminology throughout this article.
Some terms are taken over or refined from other authors and
some are new:
\begin{itemize}\parskip=0ex\parsep=0ex\itemsep=0ex
\item comp = computational resources
\item singularity = infinite change of an observable quantity in finite time
\item intelligence explosion = rapidly increasing intelligence far beyond human level
\item intelligence singularity = infinite intelligence in finite time
\item speed explosion/singularity = rapid/infinite increase of computational resources
\item outsider = biological = non-accelerated real human watching a singularity
\item insider = virtual = software intelligence participating in a singularity
\item computronium = theoretically best possible computer per unit of matter  \cite{Bremermann:65}
\item real/true intelligence = what we intuitively would regard as intelligence
\item numerical intelligence = numerical measure of intelligence like IQ score
\item AI = artificial intelligence (used generically in different ways)
\item AGI = artificial general intelligence = general human-level intelligence or beyond.
\item super-intelligence = AI+ = super-human intelligence \cite{Chalmers:10sing}
\item hyper-intelligent = AI++ = incomprehensibly more intelligent than humans
\item vorld = virtual world. A popular oxymoron is `virtual reality'
\item virtual = software simulation in a computer.
\end{itemize}
I drop the qualifier `virtual' if this does not cause any
confusion, e.g.\ when talking about a human in a vorld, I mean
of course a virtual human.

\paradot{Global assumption}
I will assume a strong/physical form of the Church-Turing
thesis that everything in nature can be calculated by a Turing
machine, i.e.\ our world including the human mind and body and
our environment are computable \cite{Deutsch:97,Hutter:11uiphil}.
So in the following I will assume without further argument that
all physical processes we desire to virtualize are indeed
computational and can be simulated by a sufficiently powerful
(theoretical) computer.
This assumption simplifies many of the considerations to
follow, but is seldom essential, and could be lifted or
weakened.

\section{Will there be a Singularity}\label{sec:WS}

\paradot{Super-intelligence by Moore's law}
The current generations Y or Z may finally realize the age-old
dream of creating systems with human-level intelligence or
beyond, which revived the interest in this endeavor. This
optimism is based on the belief that in 20--30 years the raw
computing power of a single computer will reach that of a human
brain and that software will not lag far behind.
This prediction is based on extrapolating Moore's law, now
valid for 50 years, which implies that comp doubles every 1.5
years. As long as there is demand for more comp, Moore's law
could continue to hold for many more decades before
computronium is reached.
Further, different estimates of the computational capacity of a
human brain consistently point towards $10^{15}$...$10^{16}$
flop/s \cite{Kurzweil:05}: Counting of neurons and synapses,
extrapolating tiny-brain-part simulations, and comparing the
speech recognition capacities of computers to the auditory
cortex.

\paradot{Singularity by Solomonoff's law}
The most compelling argument for the emergence of a singularity
is based on Solomonoff's law \cite{Solomonoff:85} which
Yudkowski \cite{Yudkowsky:96} succinctly describes as follows:
\begin{quote}
$\!\!\!$``If computing speeds double every two years,\\
what happens when computer-based AIs are doing the research?\\
Computing speed doubles every two years.\\
Computing speed doubles every two years of work.\\
Computing speed doubles every two subjective years of work.
\\
Two years after Artificial Intelligences reach human
equivalence, their speed doubles. One year later, their speed
doubles again.
\\
Six months - three months - 1.5 months ... Singularity.''
\end{quote}
Interestingly, if this argument is valid, then Moore's law in a
sense predicts its own break-down; not the usually anticipated
slow-down, but an enormous acceleration of progress when
measured in physical time.

\paradot{Historic acceleration of doubling patterns}
The above acceleration would indeed not be the first time of an
enormous acceleration in growth. The economist Robin Hanson
argues that ``Dramatic changes in the rate of economic growth
have occurred in the past because of some technological
advancement. Based on population growth, the economy doubled
every 250'000 years from the Paleolithic era until the
Neolithic Revolution. This new agricultural economy began to
double every 900 years, a remarkable increase. In the current
era, beginning with the Industrial Revolution, the world's
economic output doubles every fifteen years, sixty times faster
than during the agricultural era.'' Given the increasing role
of computers in our economy, computers might soon dominate it,
locking the economic growth pattern to computing speed, which
would lead to a doubling of the economy every two (or more
precisely 1.5) years, another 10 fold increase. If the rise of
superhuman intelligences causes a similar revolution, argues
Hanson \cite{Hanson:08}, one could expect the virtual economy
to double on a monthly or possibly on a weekly basis. So the
technological singularity phenomenon would be the next and
possibly last growth acceleration.
Ray Kurzweil is a master of producing exponential, double
exponential, and singular plots \cite{Kurzweil:05}, but one has
to be wary of data selection, as Juergen Schmidhuber has
pointed out.

\paradot{Obstacles towards a singularity}
Chalmers \cite{Chalmers:10sing} discusses various potential
obstacles for a singularity to emerge. He classifies them into
structural obstacles (limits in intelligence space, failure to
takeoff, diminishing returns, local maxima) and manifestation
obstacles (disasters, disinclination, active prevention) and
correlation obstacles.
For instance, self-destruction or a natural catastrophe
might wipe out the human race \cite{Bostrom:08}.

Also, the laws of physics will likely prevent a singularity in
the strict mathematical sense. While some physical theories in
isolation allow infinite computation in finite time (see Zeno
machines \cite{Weyl:27} and hypercomputation \cite{Copeland:02}
in general), modern physics raises severe barriers
\cite{Bremermann:65,Bekenstein:03,Lloyd:00,Aaronson:05comp}.
But even if so, today's computers are so far away from these
limits, that converting our planet into computronium would
still result in a vastly different vorld, which is considered a
reasonable approximation to a true singularity. Of course,
engineering difficulties and many other obstructions may stop
the process well before this point, in which case the end
result may not account as a singularity but more as a phase
transition \`a la Hanson or even less spectacular.

Like Chalmers, I also believe that disinclination is the most
(but not very) likely defeater of a singularity. In the
remainder of this article I will assume absence of any such
defeaters, and will only discuss the structural obstacles
related to limits in intelligence space later.

\paradot{Is the singularity negotiable}
The appearance of the first super-intelligences is usually
regarded as the ignition of the detonation cord towards the
singularity -- the point of no return. But it might well be
that a singularity is already now unavoidable. Politically it
is very difficult (but not impossible) to resist technology or
market forces as e.g.\ the dragging discussions on climate
change vividly demonstrate, so it would be similarly difficult
to prevent AGI research and even more so to prevent the
development of faster computers. Whether we are before, at, or
beyond the point of no return is also philosophically intricate
as it depends on how much free will one attributes to people
and society;
like a spaceship close to the event horizon might in principle
escape a black hole but is doomed in practice due to limited
propulsion.

\section{The Singularity from the Outside}\label{sec:SFO}

Let us first view the singularity from the outside. What will
observers who do not participate in it ``see''. How will it
affect them?

\paradot{Converting matter into computers}
First, the hardware (computers) for increasing comp must be
manufactured somehow. As already today, this will be done by
(real) machines/robots in factories. Insiders will provide
blue-prints to produce better computers and better machines
that themselves produce better computers and better machines ad
infinitum at an accelerated pace. Later I will explain why
insiders desire more comp. Non-accelerated real human
(outsiders) will play a diminishing role in this process due to
their cognitive and speed limitations. Quickly they will only
be able to passively observe some massive but incomprehensible
transformation of matter going on.

\paradot{Inward explosion}
Imagine an inward explosion, where a fixed amount of matter is
transformed into increasingly efficient computers until it
becomes computronium. The virtual society like a
well-functioning real society will likely evolve and progress,
or at least change. Soon the speed of their affairs will make
them beyond comprehension for the outsiders. For a while,
outsiders may be able to make records and analyze them in slow
motion with an increasing lag. Ultimately the outsiders'
recording technology will not be sufficient anymore, but some
coarse statistical or thermodynamical properties could still be
monitored, which besides other things may indicate an upcoming
physical singularity. I doubt that the outsiders will be able
to link what is going on with intelligence or a technological
singularity anymore.

Insiders may decide to interact with outsiders in slow motion
and feed them with pieces of information at the maximal
digestible rate, but even with direct brain-computer
interfaces, the cognitive capacity of a human brain is bounded
and cannot explode. A technologically augmented brain may
explode, but what would explode is the increasingly dominant
artificial part, rendering the biological brain eventually
superfluous --- a gradual way of getting sucked into the inside
world. For this reason, also intelligence amplification by
human-computer interfaces are only temporarily viable before
they either break down or the extended human becomes
effectively virtual.

After a brief period, intelligent interaction between insiders
and outsiders becomes impossible. The inside process may from
the outside resemble a black hole watched from a safe distance,
and look like another interesting physical, rather than
societal, phenomenon.

\paradot{Information}
This non-comprehensibility conclusion can be supported by an
information-theoretic argument: The characterization of our
society as an information society becomes even better, if not
perfect, for a virtual society. There is lots of motivation to
compress information (save memory, extract regularities, and
others), but it is well-known \cite{Li:08} that maximally
compressed information is indistinguishable from random noise.
Also, if too much information is produced, it may actually
``collapse''. Here, I am not referring to the formation of
black holes \cite{Bekenstein:03}, but to the fact that a
library that contains all possible books has zero information
content (cf.\ the Library of Babel). Maybe a society of
increasing intelligence will become increasingly
indistinguishable from noise when viewed from the outside.

\paradot{Outward explosion}
Let us now consider outward explosion, where an increasing
amount of matter is transformed into computers of fixed
efficiency (fixed comp per unit time/space/energy). Outsiders
will soon get into resource competition with the expanding
computer world, and being inferior to the virtual
intelligences, probably only have the option to flee. This
might work for a while, but soon the expansion rate of the
virtual world should become so large, theoretically only
bounded by the speed of light, that escape becomes impossible,
ending or converting the outsiders' existence.

\paradot{Comparison}
So while an inward explosion is interesting, an outward
explosion will be a threat to outsiders. In both cases,
outsiders will observe a speedup of cognitive processes and
possibly an increase of intelligence up to a certain point. In
neither case will outsiders be able to witness a true
intelligence singularity.

\paradot{Historical human expansion/exploration analogy}
Historically, mankind was always outward exploring; just in
recent times it has become more inward exploring. Now people
more and more explore virtual worlds rather than new real
worlds. There are two reasons for this. First, virtual worlds
can be designed as one sees fit and hence are arguably more
interesting, and second, outward expansion now means deep sea
or space, which is an expensive endeavor. Expansion usually
follows the way of least resistance.

\paradot{Path of least resistance}
Currently the technological explosion is both inward and
outward (more and faster computers). Their relative speed in
the future will depend on external constraints. Inward
explosion will stop when computronium is reached. Outward
explosion will stop when all accessible convertible matter has
been used up (all on earth, or in our galaxy, or in our
universe).

\section{The Singularity from the Inside}\label{sec:SFI}

Let us now consider the singularity from the inside.
What will a participant experience?

\paradot{Virtualize society}
Many things of course will depend on how the virtual world is
organized. It is plausible that various characteristics of our
current society will be incorporated, at least initially. Our
world consists of a very large number of individuals, who
possess some autonomy and freedom, and who interact with each
other and with their environment in cooperation and in
competition over resources and other things.
Let us assume a similar setup in a virtual world of intelligent
actors. The vorld might actually be quite close to our real
world. Imagine populating already existing virtual worlds like
Second Life or World of Warcraft with intelligent agents
simulating scans of human brains.

\paradot{Fixed comp}
Consider first a vorld based on fixed computational resources.
As indicated, initially, the virtual society might be similar
to its real counter-part, if broadly understood. But some
things will be easier, such a duplicating (virtual) objects and
directed artificial evolution. Other things will be harder or
impossible, such as building faster virtual computers and
fancier gadgets reliant on them.
This will affect how the virtual society will value different
things (the value of virtual life and its implications will be
discussed later), but I would classify most of this as a
change, not unlike in the real world when discovering or
running out of some natural resource or adapting to new models
of society and politics.
Of course, the virtual society, like our real one, will also
develop: there will be new inventions, technologies, fashions,
interests, art, etc., all virtual, all software, of course, but
for the virtuals it will feel real.
If virtuals are isolated from the outside world and have
knowledge of their underlying computational processes, there
would be no quest for a virtual theory of everything
\cite{Hutter:10ctoex}, since they would already know it.
The evolution of this vorld might include weak singularities in
the sense of sudden phase transitions or collapses of the
society, but an intelligence explosion with fixed comp, even
with algorithmic improvements seems implausible.

\paradot{Increasing comp (per individual)}
Consider now the case of a vorld with increasing comp. If extra
comp is used for speeding up the whole virtual world uniformly,
virtuals and their virtual environment alike, the inhabitants
would actually not be able to recognize this. If their
subjective thought processes will be sped up at the same rate
as their surroundings, nothing would change for them. The only
difference, provided virtuals have a window to the outside real
world, would be that the outside world slows down. If comp is
sped up hyperbolically, the subjectively infinite future of the
virtuals would fit into finite real time: For the virtuals, the
external universe would get slower and slower and ultimately
come to a halt. Also outsiders would appear slower (but not
dumber).

\paradot{Black hole versus technological singularity}
This speed-up/slow-down phenomenon is inverse compared to
flying into a black hole. An astronaut flying into a black hole
will pass the Schwarzschild radius and hit the singularity in
finite subjective time. For an outside observer, though, the
astronaut gets slower and slower and actually takes infinite
time to vanish behind the Schwarzschild radius.

\paradot{Increasing comp (number of individuals)}
If extra comp is exclusively used to expand the vorld and add
more virtuals, there is no individual speedup, and the bounded
individual comp forces intelligence to stay bounded, even with
algorithmic improvements. But larger societies can also evolve
faster (more inventions per real time unit), and if regarded as
a super-organism, there might be an intelligence explosion, but
not necessarily so: Ant colonies and bee hives seem more
intelligent than their individuals in isolation, but it is not
obvious how this scales to unbounded size. Also, there seems to
be no clear positive correlation between the number of
individuals involved in a decision process and the intelligence
of its outcome.

\paradot{No intelligence explosion}
In any case, the virtuals as individuals will not experience an
intelligence explosion, even if there was one. The outsiders
would observe virtuals speeding up beyond comprehension and
would ultimately not recognize any further intelligence
explosion.

The scenarios considered in this and the last section are of
course only caricatures. An actual vorld will more likely
consist of a wide diversity of intelligences: faster and slower
ones, higher and lower ones, and a hierarchy of super-organisms
and sub-vorlds. The analysis becomes more complicated, but the
fundamental conclusion that an intelligence explosion might be
unobservable does not change.

\section{Speed versus Intelligence Explosion}\label{sec:SVIE}

The comparison of the inside and outside view has revealed that
a speed explosion is not necessarily an intelligence explosion.
In the extreme case, insiders may not experience anything and
outsiders may witness only noise.

\paradot{Agent-environment model}
Consider an agent interacting with an environment. If both are
sped up at the same rate, their behavioral interaction will not
change except for speed. If there is no external clock
measuring absolute time, there is no net effect at all.

\paradot{Speed up environment}
If only the environment is sped up, this has the same effect as
slowing down the agent. This does not necessarily make the
agent dumber. He will receive more information per action, and
can make more informed decisions, provided he is left with
enough comp to process the information. Imagine being inhibited
by very slowly responding colleagues. If you could speed them
up, this would improve your own throughput, and subjectively
this is the same as slowing yourself down. But (how much) can
this improve the agent's intelligence? In the extreme case,
assume the agent has instant access to all information, not
much unlike we already have by means of the Internet but much
faster. Both usually increase the quality of decisions, which
might be viewed as an increase in intelligence. But intuitively
there should be a limit on how much information a comp-limited
agent can usefully process or even search through.

\paradot{Speed up agent}
Consider now the converse and speed up the agent (or
equivalently slow down the environment). From the agent's view,
he becomes deprived of information, but has now increased
capacity to process and think about his observations. He
becomes more reflective and cognitive, a key aspect of
intelligence, and this should lead to better decisions. But
also in this case, although it is much less clear, there might
be a limit to how much can be done with a limited amount of
information.

\paradot{Speed summary}
The speed-up/slow-down effects might be summarized as follows:

\begin{itemize}\parskip=0ex\parsep=0ex\itemsep=0ex
\item[] $\!\!\nq\nq$ Performance per unit real time:
\item Speed of agent positively correlates with cognition and intelligence of decisions
\item Speed of environment positively correlates with informed decisions
\item[] $\nq\nq$Performance per subjective unit of agent time from agent's perspective:
\item slow down environment = increases cognition and intelligence but decisions become less informed
\item speed up environment = more informed but less reasoned decisions
\item[] $\nq\nq$Performance per environment time from environment perspective:
\item speed up agent = more intelligent decisions
\item slow down agent = less intelligent decisions
\end{itemize}

\paradot{Some more thoughts on speed}
I have argued that more comp, i.e.\ speeding up hardware, does
not necessarily correspond to more intelligence. But then the
same could be said of software speedups, i.e.\ more efficient
ways of computing the same function. If two agent algorithms
have the same I/O behavior, just one is faster than the other,
is the faster one more intelligent?

An interesting related question is whether progress in AI has
been mainly due to improved hardware or improved software. If
we believe in the former, and we accept that speed is
orthogonal to intelligence, and we believe that humans are
``truly'' intelligent (a lot of ifs), then building AGIs may
still be far distant.

As detailed in Section~\ref{sec:IUB}, if intelligence is
upper-bounded (like playing optimal minimax chess), then past
this bound, intelligences can only differ by speed and
available information to process. In this case, and if humans
are not too far below this upper bound (which seems unlikely),
outsiders could, as long as their technology permits, record
and play a virtual world in slow motion and be able to grasp
what is going on inside.

In this sense, a singularity may be more interesting for
outsiders than for insiders. On the other hand, insiders
actively ``live'' potential societal changes, while outsiders
only passively observe them.

Of course, more comp only leads to more intelligent decisions
if the decision algorithm puts it to good use. Many algorithms
in AI are so-called anytime algorithms that indeed produce
better results if given more comp. In the limit of infinite
comp, in simple and well-defined settings (usually search and
planning problems), some algorithms can produce optimal
results, but for more realistic complex situations (usually
learning problems), they saturate and remain sub-optimal
\cite{Russell:10}. But there is one algorithm, namely AIXI
described in Section~\ref{sec:IUB}, that is able to make
optimal decisions in arbitrary situations given infinite comp.

Together this shows that it is non-trivial to draw a clear
boundary between speed and intelligence.

\section{What is Intelligence}\label{sec:WII}

There have been numerous attempts to define intelligence; see
e.g.\ \cite{Hutter:07idefs} for a collection of 70+ definitions
from the philosophy, psychology, and AI literature, by
individual researchers as well as collective attempts.

If/since intelligence is not (just) speed, what is it then?
What will super-intelligences actually do?

\paradot{Historical/biological/evolutionary/memetic intelligence}
Historically-biologically, higher intelligence, via some
correlated practical cognitive capacity, increased the chance
of survival and number of offspring of an individual and the
success of a species. At least for primates leading to homo
sapiens this was the case until recently. Within the human
race, intelligence is now positively correlated with power
and/or economic success \cite{Geary:07} and actually negatively
with number of children \cite{Kanazawa:07}. Genetic evolution
has been largely replaced by memetic evolution
\cite{Dawkins:76}, the replication, variation, selection, and
spreading of ideas causing cultural evolution.

\paradot{What activities are intelligent}
What activities could be regarded as or are positively
correlated with intelligence?
Self-preservation? %
Self-replication? Spreading? %
Creating faster/better/higher intelligences? %
Learning as much as possible? %
Understanding the universe? %
Maximizing power over men and/or organizations? %
Transformation of matter (into computronium?)? %
Maximum self-sufficiency?
The search for the meaning of life? %

\paradot{Intelligence = rationality = reasoning towards a goal}
Has intelligence more to do with thinking or is thinking only a
tool for acting smartly? Is intelligence something
anthropocentric or does it exist objectively?
What are the relations between other predicates of human
``spirit'' like consciousness, emotions, and religious faith to
intelligence? Are they part of it or separate characteristics
and how are they interlinked?

One might equate intelligence with rationality, but what is
rationality? Reasoning, which requires internal logical
consistency, is a good start for a characterization but is
alone not sufficient as a definition. Indiscriminately
producing one true statement after the other without
prioritization or ever doing anything with them is not too
intelligent (current automated theorem provers can already do
this).

It seems hard if not impossible to define rationality without
the notion of a goal. If rationality is reasoning towards a
goal, then there is no intelligence without goals. This idea
dates back at least to Aristotle, if not further; see
\cite{Hutter:07iorx} for details. But what are the goals?
Slightly more flexible notions are that of expected utility
maximization and cumulative life-time reward maximization
\cite{Russell:10}. But who provides the rewards, and how? For
animals, one might try to equate the positive and negative
rewards with pleasure and pain, and indeed one can explain a
lot of behavior as attempts to maximize rewards/pleasure.
Humans seem to exhibit astonishing flexibility in choosing
their goals and passions, especially during childhood.
Goal-oriented behavior often appears to be at odds with
long-term pleasure maximization. Still, the evolved biological
goals and desires to survive, procreate, parent, spread,
dominate, etc.\ are seldom disowned.

\paradot{Evolving goals}
But who sets the goal for super-intelligences and how? When
building AIs or tinkering with our virtual selves, we could try
out a lot of different goals, e.g.\ selected from the list
above or others. But ultimately we will lose control, and the
AGIs themselves will build further AGIs (if they were motivated
to do so) and this will gain its own dynamic. Some aspects of
this might be independent of the initial goal structure and
predictable. Probably this initial vorld is a society of
cooperating and competing agents. There will be competition
over limited (computational) resources, and those virtuals who
have the goal to acquire them will naturally be more successful
in this endeavor compared to those with different goals.
Of course, improving the efficiency of resource use is
important too, e.g.\ optimizing own algorithms, but still,
having more resources is advantageous.
The successful virtuals will spread (in various ways), the
others perish, and soon their society will consist mainly of
virtuals whose goal is to compete over resources, where
hostility will only be limited if this is in the virtuals' best
interest. For instance, current society has replaced war mostly
by economic competition, since modern weaponry makes most wars
a loss for both sides, while economic competition in most cases
benefits the better.

\paradot{The goal to survive and spread}
Whatever amount of resources are available, they will (quickly)
be used up, and become scarce. So in any world inhabited by
multiple individuals, evolutionary and/or economic-like forces
will ``breed'' virtuals with the goal to acquire as much (comp)
resources as possible. This world will likely neither be heaven
nor hell for the virtuals. They will ``like'' to fight over
resources, and the winners will ``enjoy'' it, while the losers
will ``hate'' it. In such evolutionary vorlds, the ability to
survive and replicate is a key trait of intelligence. On the
other hand, this is not a sufficient characterization, since
e.g.\ bacteria are quite successful in this endeavor too, but
not very intelligent.

\paradot{Alternative societies}
Finally, let us consider some alternative (real or virtual)
worlds.
In the human world, local conflicts and global war is
increasingly replaced by economic competition, which might
itself be replaced by even more constructive global
collaboration, as long as violaters can quickly and effectively
(and non-violently?) be eliminated.
It is possible that this requires a powerful single (virtual)
world government, to give up individual privacy, and to
severely limit individual freedom (cf.\ ant hills or bee hives).
An alternative societal setup that can only produce conforming
individuals might only be possible by severely limiting
individual's creativity (cf.\ flock of sheep or school of fish).

Such well-regulated societies might better be viewed as a
single organism or collective mind. Or maybe the vorld is
inhabited from the outset by a single individual.
Both vorlds could look quite different and more peaceful than
the traditional ones created by evolution.
Intelligence would have to be defined quite differently in such
vorlds.
Many science fiction authors have conceived and extensively
written about a plethora of other future, robot, virtual, and
alien societies in the last century.

In the following I will only consider vorlds shaped by
evolutionary pressures as described above.

\section{Is Intelligence Unlimited or Bounded}\label{sec:IUB}

Another important aspect of intelligence is how flexible or
adaptive an individual is. Deep blue might be the best chess
player on Earth, but is unable to do anything else. On the
contrary, higher animals and humans have remarkably broad
capacities and can perform well in a wide range of
environments.

\paradot{Formal intelligence measure}
In \cite{Hutter:07iorx} intelligence has been defined as the
ability to achieve goals in a wide range of environments. It
has been argued that this is a very suitable characterization,
implicitly capturing most, if not all traits of rational
intelligence, such as reasoning, creativity, generalization,
pattern recognition, problem solving, memorization, planning,
learning, self-preservation, and many others. Furthermore, this
definition has been rigorously formalized in mathematical
terms. It is non-anthropocentric, wide-range, general,
unbiased, fundamental, objective, complete, and universal. It
is the most comprehensive formal definition of intelligence so
far. It assigns a real number $\Upsilon$ between zero and one
to every agent, namely the to-be-expected performance averaged
over all environments/problems the agent potentially has to
deal with, with an Ockham's razor inspired prior weight for
each environment. Furthermore there is a maximally intelligent
agent, called AIXI, w.r.t.\ this measure. The precise formal
definitions and details can be found in \cite{Hutter:07iorx},
but do not matter for our purpose. This paper also contains a
comprehensive justification and defense of this approach.

The theory suggests that there is a maximally intelligent
agent, or in other words, that intelligence is upper bounded
(and is actually lower bounded too). At face value, this would
make an intelligence explosion impossible.

\paradot{Motivation: tic-tac-toe and chess}
To motivate this possibility, consider some simple examples.
Assume the vorld consists only of tic-tac-toe games, and the
goal is to win or second-best not lose them. The notion of
intelligence in this simple vorld is beyond dispute. Clearly
there is an optimal strategy (actually many) and it is
impossible to behave more intelligently than this strategy. It
is even easy to artificially evolve or learn these strategies
from repeated (self)play \cite{Hochmuth:03,Hutter:11aixictwx}.
So in this vorld there clearly will be no intelligence
explosion or intelligence singularity, even if there were a
speed explosion.

We get a slightly different situation when we replace
tic-tac-toe by chess. There is also an optimal way of playing
chess, namely minimax tree search to the end of the game, but
unlike in tic-tac-toe this strategy is computationally
infeasible in our universe. So in theory (i.e.\ given enough
comp) intelligence is upper-bounded in a chess vorld, while in
practice we can get only ever closer but never reach the bound.
(Actually there might be enough matter in the universe to build
an optimal chess player, but likely not an optimal Go player.
In any case it is easy to design a game that is beyond the
capacity of our accessible universe, even if completely
converted into computronium).

Still, this causes two potential obstacles for an intelligence
explosion. First, we are only talking about the speed of
algorithms, which I explained before not to equate with
intelligence. Second, intelligence is upper bounded by the
theoretical optimal chess strategy, which makes an intelligence
explosion difficult but not necessarily impossible: Assume the
optimal program has intelligence $I=1$ and at real time $t<1$
we have access to or evolved a chess program with intelligence
$t$. This approaches 1 in finite time, but doesn't ``explode''.
But if we use the monotone transformation $1/(1-I)$ to measure
intelligence, the chess program at time $t$ has transformed
intelligence $1/(1-t)$ which tends to infinity for $t\to 1$.
While this is a mathematical singularity, it is likely not
accompanied by a real intelligence explosion. The original
scale seems more plausible in the sense that $t+0.001$ is just
a tiny bit more intelligent than $t$, and $1$ is just 1000
times more intelligent than $0.001$ but not infinitely more.
Although the vorld of chess is quite rich, the real world is
vastly and possibly unlimitedly richer. In such a more open
world, the intelligence scale may be genuinely unbounded, but
not necessarily as we will see.
It is not easy though to make these arguments rigorous.

\paradot{Real world and AIXI}
Let us return to the real world and intelligent measure
$\Upsilon$ upper bounded by $\Upsilon_{max}=\Upsilon$(AIXI).
Since AIXI is incomputable, we can never reach intelligence
$\Upsilon_{max}$ in a computational universe, but similarly to
the chess example we can get closer and closer. The numerical
advance is bounded, and so is possibly the real intelligence
increase, hence no intelligence explosion. But it might also be
the case that in a highly sophisticated AIXI-close society, one
agent beating another by a tiny epsilon on the $\Upsilon$-scale
makes all the difference for survival and/or power and/or other
measurable impact like transforming the universe. In many sport
contests split seconds determine a win, and the winner takes it
all --- an admittedly weak analogy.

\paradot{Intelligence of current humans}
An interesting question is where humans range on the
$\Upsilon$-scale: is it so low with so much room above that
outsiders would effectively experience an intelligence
explosion (as far as recognizable), even if intelligence is
ultimately upper bounded? Or are we already quite close to the
upper bound, so that even AGIs with enormous comp (but
comparable I/O limitations) would just be more intelligent but
not incomprehensibly so. We tend to believe that we are quite
far from $\Upsilon$, but is this really so? For instance, what
has once been argued to be irrational (i.e. not very
intelligent) behavior in the past, can often be regarded as
rational w.r.t.\ the appropriate goal. Maybe we are already
near-optimal goal achievers. I doubt this, but cannot rule it
out either.

\paradot{Extrapolation}
Humans are not faster but more intelligent than dogs, and dogs
in turn are more intelligent than worms and not just faster,
even if we cannot pinpoint exactly why we are more intelligent:
is it our capacity to produce technology or to transform our
environment on a large scale or consciousness or domination
over all other species? There are no good arguments why humans
should be close to the top of the possible biological
intelligence scale, and even less so on a vorld scale. By
extrapolation it is plausible that a vorld of much more
intelligent trans-humans or machines is possible. They will
likely be able to perform better in an even wider range of
environments on an even wider range of problems than humans.
Whether this results in anything that deserves the name
intelligence explosion is unclear.

\section{Singularitarian Intelligences}\label{sec:SI}

\paradot{Singularity = society of AIXIs}
Consider a vorld inhabited by competing agents, initialized
with human mind-uploads or non-human AGIs, and increasing comp
per virtual.
Sections~\ref{sec:WII} and~\ref{sec:IUB} then indicate that
evolutionary pressure increases the individuals' intelligence
and the vorld should converge to a society of AIXIs.
Alternatively, if we postulate an intelligence singularity and
accept that AIXI is the most intelligent agent, we arrive at
the same conclusion. More precisely, the society consists of
agents that aim at being AIXIs only being constrained by comp.
If this is so, the intelligence singularity might be identified
with a society of AIXIs, so studying AIXI can tell us something
about how a singularity might look like. Since AIXI is
completely and formally defined, properties of this society can
be studied rigorously mathematically. Here are some questions
that could be asked and answered:

\paradot{Social questions regarding a society of AIXIs}
Will a pure reward maximizer such as AIXI listen to and trust a teacher? Likely yes. %
Will it take drugs (i.e.\ hack the reward system)? Likely no, since cumulative long-term reward would be small (death). %
Will AIXI replicate itself or procreate? Likely yes, if AIXI believes that clones or descendants are useful for its own goals. %
Will AIXI commit suicide? Likely yes (no), if AIXI is raised to believe in going to heaven (hell) i.e.\ maximal (minimal) reward forever. %
Will sub-AIXIs self-improve? Likely yes, since this helps to increase reward. %
Will AIXI manipulate or threaten teachers to give more reward? Likely yes. %
Are pure reward maximizers like AIXI egoists, psychopaths, and/or killers, or will they be friendly (altruism as extended ego(t)ism)? %
Curiosity killed the cat and maybe AIXI, or is extra reward for curiosity necessary? %
Immortality can cause laziness. Will AIXI be lazy? %
Can self-preservation be learned or need (parts of) it be innate. %
How will AIXIs interact/socialize in general? %

\paradot{On answering questions regarding a society of AIXIs}
For some of these questions, partial and informal discussions
and plausible answers are available, and a couple have been
rigorously defined, studied and answered, but most of them are open to
date
\cite{Hutter:04uaibook,Schmidhuber:07curiosity,Orseau:11agi,Ring:11,Hutter:12uaigentle}.
But the AIXI theory has the potential to arrive at definite answers
to various questions regarding the social behavior of
super-intelligences close to or at an intelligence singularity.

\section{Diversity Explosion and the Value of a Virtual Life}\label{sec:DE}\label{sec:VVL}

\paradot{Copying and modifying virtual structures}
As indicated, some things will be harder or impossible in a
virtual world (e.g.\ to discover new physics) but many things
should be easier. Unless a global copy protection mechanism is
deliberately installed (like e.g.\ in Second Life) or copyright
laws prevent it, copying virtual structures should be as cheap
and effortless as it is for software and data today. The only
cost is developing the structures in the first place, and the
memory to store and the comp to run them. With this comes the
possibility of cheap manipulation and experimentation.

\paradot{Copying and modifying virtual life}
It becomes particularly interesting when virtual life itself
gets copied and/or modified. Many science fiction stories cover
this subject, so I will be brief and selective here.
One consequence should be a ``virtuan'' explosion with life
becoming much more diverse. Andy Clarke \cite{Clark:09} writes
(without particularly referring to virtuals) that ``The humans
of the next century will be vastly more heterogenous, more
varied along physical and cognitive dimensions, than those of
the past as we deliberately engineer a new Cambrian explosion
of body and mind.'' In addition, virtual lives could be
simulated in different speeds, with speeders experiencing
slower societal progress than laggards. Designed intelligences
will fill economic niches. Our current society already relies
on specialists with many years of training, so it is natural to
go the next step to ease this process with ``designer babies''.

\paradot{The value of life}
Another consequence should be that life becomes less valuable.
Our society values life, since life is a valuable commodity and
expensive/laborious to replace/produce/raise. We value our own
life, since evolution selects only organisms that value their
life. Our human moral code mainly mimics this, with cultural
differences and some excesses (e.g.\ suicide attacks on the one
side and banning stem cell research on the other).

If life becomes `cheap', motivation to value it will decline.
Analogies are abundant: Cheap machines decreased the value of
physical labor. Some expert knowledge was replaced by
hand-written documents, then printed books, and finally
electronic files, where each transition reduced the value of
the same information. Digital computers made human computers
obsolete. In games, we value our own life and that of our
opponents less than real life, not only because a game is a
crude approximation to real life, but also because games can be
reset and one can be resurrected. Governments will stop paying
my salary when they can get the same research output from a
digital version of me, essentially for free.

And why not participate in a dangerous fun activity if in the
worst case I have to activate a backup copy of myself from
yesterday which just missed out this one (anyway not too
well-going) day. The belief in immortality can alter behavior
drastically.

\paradot{The value of virtual life}
Of course there will be countless other implications:
ethical, political, economical, medical, cultural,
humanitarian, religious, in art, warfare, etc.
I have singled out the value of life, since I think it will
significantly influence other aspects. Much of our society is
driven by the fact that we highly value (human/individual)
life. If virtual life is/becomes cheap, these drives will
ultimately vanish and be replaced by other goals.
If AIs can be easily created, the value of an intelligent
individual will be much lower than the value of a human life
today. So it may be ethically acceptable to freeze, duplicate,
slow-down, modify (brain experiments), or even kill (oneself or
other) AIs at will, if they are abundant and/or backups are
available, just what we are used to doing with software. So
laws preventing experimentation with intelligences for moral
reasons may not emerge. With so little value assigned to an
individual life, maybe it becomes a disposable.

\section{Personal Remarks}\label{sec:Misc}

\paradot{Consciousness}
I have deliberately avoided discussing consciousness for
several reasons: David Chalmers is {\em the} consciousness
expert and not me, he has extensively written about it in
general and also in the context of the singularity
\cite{Chalmers:10sing}, and I essentially agree with his
assessments. Personally I believe in the functionalist theory
of identity and am confident that (slow and fast) uploading of
a human mind preserves identity and consciousness, and indeed
that any sufficiently high intelligence, whether
real/biological/physical or virtual/silicon/software is
conscious, and that consciousness survives changes of
substrate: teleportation, duplication, virtualization/scanning,
etc.\ along the lines of \cite{Chalmers:10sing}.

\paradot{Desirable futures}
I have also only considered (arguably) plausible scenarios, but
not whether these or other futures are desirable. First, there
is the problem of how much influence/choice/freedom we actually
have in shaping our future in general and the singularity in
particular. Can evolutionary forces be beaten?
Second, what is desirable is necessarily subjective. Are there
any universal values or qualities we want to see or that should
survive? What do I mean by we? All humans? Or the dominant
species or government at the time the question is asked? Could
it be diversity? Or friendly AI \cite{Yudkowsky:08}? Could the
long-term survival of at least one conscious species that
appreciates its surrounding universe be a universal value? A
discussion of these questions is clearly beyond the scope of
this article.

\section{Conclusions}\label{sec:Conc}

Based on the deliberations in this paper, here are my
predictions concerning a potential technological singularity,
although admittedly they have a speculative character.

\begin{itemize}\parskip=0ex\parsep=0ex\itemsep=0ex
\item This century may witness a technological explosion of a
degree deserving the name singularity.

\item The default scenario is a society of interacting intelligent
agents in a virtual world, simulated on computers with
hyperbolically increasing computational resources.

\item This is inevitably accompanied by a speed explosion when
measured in physical time units, but not necessarily by an
intelligence explosion.

\item Participants will not necessarily experience this explosion,
since/if they are themselves accelerated at the same pace,
but they should enjoy `progress' at a `normal' subjective pace.

\item For non-accelerated non-participating conventional humans,
after some short period, their limited minds will not be able
to perceive the explosion as an intelligence explosion.

\item This begs the question in which sense an intelligence
explosion has happened. (If a tree falls in a forest and no one
is around to hear it, does it make a sound?)

\item One way and maybe the only way to make progress in this
question is to clarify what intelligence actually is.

\item The most suitable notion of intelligence for this purpose seems
to be that of universal intelligence, which in principle allows
to formalize and theoretically answer a wide range of questions
about super-intelligences. Accepting this notion has
in particular the following implications:

\item There is a maximally intelligent agent, which appears to imply
that intelligence is fundamentally upper bounded, but this is
not necessarily so.

\item If the virtual world is inhabited by interacting free
agents (rather than a `monistic' vorld inhabited by a
single individual or a tightly controlled society),
evolutionary pressures should breed agents of increasing
intelligence that compete about computational resources.

\item The end-point of this intelligence evolution/acceleration
(whether it deserves the name singularity or not)
could be a society of these maximally intelligent
individuals.

\item Some aspects of this singularitarian society might be
theoretically studied with current scientific tools.

\item Way before the singularity, even when setting up a virtual
society in our image, there are likely some immediate
differences, for instance that the value of an individual life
suddenly drops, with drastic consequences.
\end{itemize}

\noindent{\bf Acknowledgements.}
Thanks to Wolfgang Schwarz and Reinhard Hutter
for feedback on earlier drafts.

\addcontentsline{toc}{section}{\refname}
\bibliographystyle{alpha}

\begin{small}
\newcommand{\etalchar}[1]{$^{#1}$}

\end{small}

\end{document}